\documentclass{article}
\usepackage{ijcai20}

\usepackage{times}
\usepackage{soul}
\usepackage{url}
\usepackage[backref=nolink]{}
\usepackage[utf8]{inputenc}
\usepackage[small]{caption}
\usepackage{graphicx}
\usepackage{amsmath}
\usepackage{amsthm}
\usepackage{booktabs}
\usepackage{algorithm}
\usepackage{algorithmic}
\usepackage{amssymb}

\begin{document}

\title{Baby Steps Towards Few-Shot Learning with Multiple Semantics}



\renewcommand\footnotemark{}
\author{Eli Schwartz*\textsuperscript{1,2}\thanks{*The authors have contributed equally to this work}, Leonid Karlinsky*\textsuperscript{1}, Rogerio Feris\textsuperscript{1}, Raja Giryes\textsuperscript{2} and Alex M. Bronstein\textsuperscript{3} \\ \ \\
\textsuperscript{1}IBM Research AI;
\textsuperscript{2}Tel-Aviv University;
\textsuperscript{3}Technion
}

\maketitle
\let\thefootnote\relax\footnotetext{

Corresponding authors:
Eli Schwartz elisch@ibm.com; Leonid Karlinsky leonidka@il.ibm.com
}

\begin{abstract}
   Learning from one or few visual examples is one of the key capabilities of humans since early infancy, but is still a significant challenge for modern AI systems. While considerable progress has been achieved in few-shot learning from a few image examples, much less attention has been given to the verbal descriptions that are usually provided to infants when they are presented with a new object. In this paper, we focus on the role of additional semantics that can significantly facilitate few-shot visual learning. Building upon recent advances in few-shot learning with additional semantic information, we demonstrate that further improvements are possible by combining multiple and richer semantics (category labels, attributes, and natural language descriptions). Using these ideas, we offer the community new results on the popular \textit{mini}ImageNet and CUB few-shot benchmarks, comparing favorably to the previous state-of-the-art results for both visual only and visual plus semantics-based approaches. We also performed an ablation study investigating the components and design choices of our approach.
\end{abstract}

\begin{figure}[htb]
  \centering
    \includegraphics[width=0.83\linewidth]{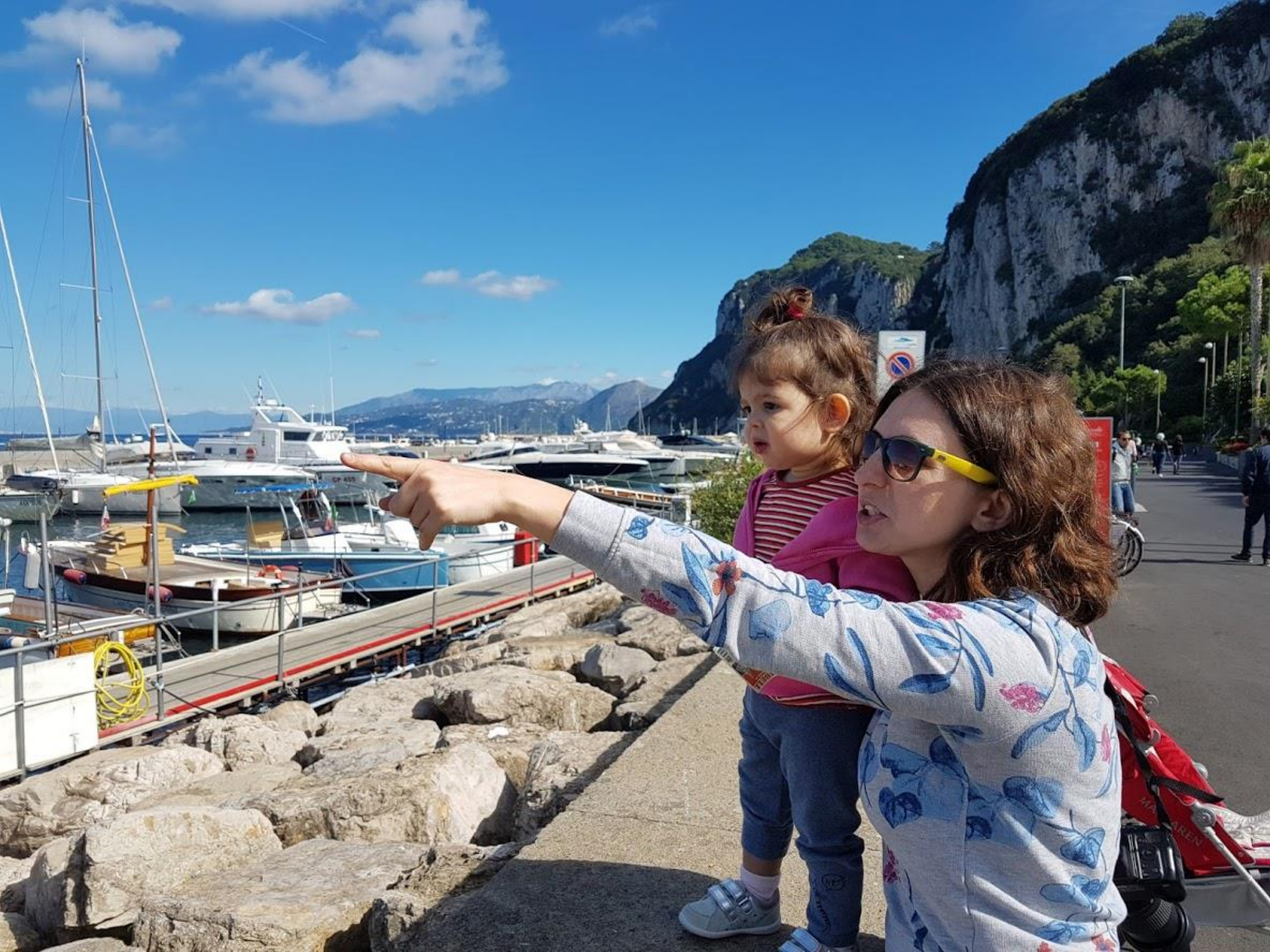}
    \caption{Pointing out to a new kind of object for a child is often accompanied by additional associated (multiple) semantic information.}
\label{fig:concept}
\end{figure}

\begin{figure*}[t]
  \centering
    \includegraphics[width=0.95\linewidth]{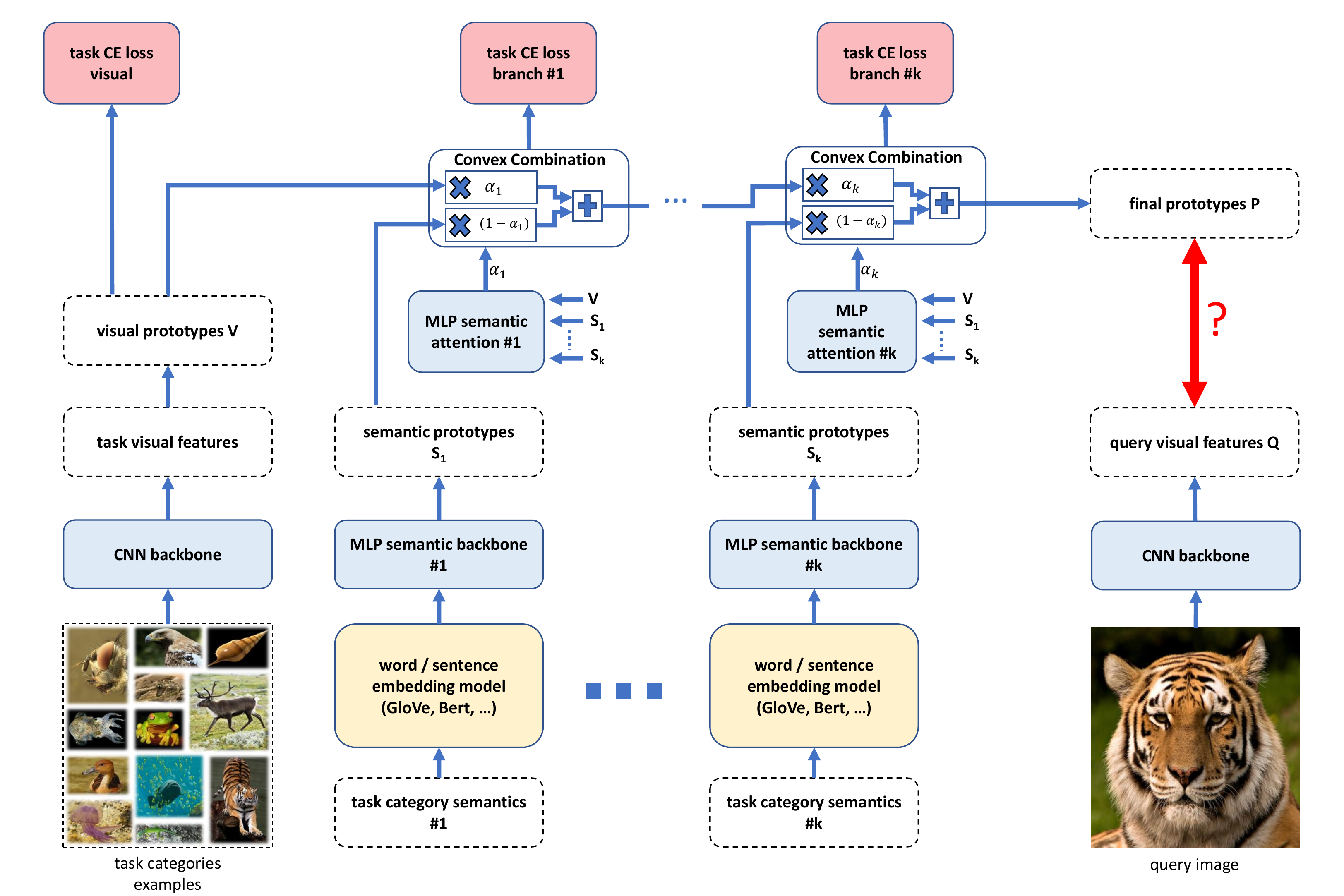}
  \caption{The proposed model, best viewed in color. Some connecting lines are excluded for brevity. Filled boxes represent neural nets and losses, bluish nets are jointly learned as part of our approach, yellowish ones are for word / sentence embedding and are pre-trained and fixed. Please see section \ref{sec:method} for details.}
\label{fig:model}
\end{figure*}

\section{Introduction}\label{sec:intro}
Modern day computer vision has experienced a tremendous leap due to the advent of deep learning (DL) techniques. The DL-based approaches reach higher levels of performance even compared to humans in tasks requiring expertise, such as recognizing dog breeds, or faces of thousands of celebrities. Yet, despite all the advances, some innate human abilities available to us at a very young age, still elude modern AI systems. One of these abilities is to be able to learn and later successfully recognize new, previously unseen, visual categories when presented to us with one or very few examples. 
This `few-shot learning' task has been thoroughly explored in the computer vision literature and numerous approaches have been proposed (please see \cite{closer_look} for a recent review). Yet so far, the performance of even the best few-shot learning methods fall short by a significant margin from the performance of the fully supervised learning methods trained with a large number of examples (for example ImageNet \cite{imagenet}).

One important ingredient of human infant learning, which has only very recently found its way into the visual few-shot learning approaches, is the associated semantics that comes with the provided example.
For example, it has been shown in the child development literature that infants' object recognition ability is linked to their language skills and it is hypothesized that it might be related to the ability to describe objects \cite{smith2003learning}.
Indeed, when a parent points a finger at a new category to be learned (`look, here is a puppy', figure \ref{fig:concept}), it is commonly accompanied by additional semantic references or descriptions for that category (e.g., `look at his nice fluffy ears', `look at his nice silky fur', `the puppy goes woof-woof'). This additional, and seldom rich, semantic information can be very useful to the learner, and has been exploited in the context of zero-shot learning and visual-semantic embeddings. Indeed, language as well as vision domains, both describe the same physical world in different ways, and in many cases contain useful complementary information that can be carried over to the learner in the other domain (visual to language and vice versa).

In the recent few-shot learning literature, the additional power of using semantics to facilitate few-shot learning was realized in only a handful of works. In \cite{Chen2018} an embedding vector of either the category label or of the given set of category attributes is used to regularize the latent representation of an auto-encoder TriNet network by adding a loss for making the sample latent vector as close as possible to the corresponding semantic vector. In \cite{am3} the semantic representation of visual categories is learned on top of the GloVe \cite{glove} word embedding, jointly with a Proto-Net \cite{Snell2017} based few-shot classifier, and jointly with the convex combination of both. The result of this joint training is a powerful few-shot and zero-shot (that is a semantic-based) ensemble that surpassed the performance of all other few-shot learning methods to-date on the challenging \textit{mini}ImageNet few-shot learning benchmark \cite{Vinyals2016}. In both of these cases, combining few-shot learning with some category semantics (labels or attributes) proved highly beneficial to the performance of the few-shot learner. Yet in both cases, only the simple one word embedding or a set of several prescribed numerical attributes were used to encode the semantics. 

In this work, we show that more can be gained by exploring a more realistic human-like learning setting. This is done by providing the learner access to multiple and richer semantics. These semantics can include, depending on what available for the dataset: category labels; richer `description level' semantic information (a sentence, or a few sentences, in a natural language with a description of the category); or attributes.
We demonstrate how this learning with semantic setting can facilitate few-shot learning (leveraging the intuition of how human infants learn). The results compare favorably to the previous visual and visual + semantics state-of-the-art results on the challenging \textit{mini}ImageNet \cite{Vinyals2016} and CUB \cite{WelinderEtal2010} few-shot benchmarks.

To summarize, the contributions of this work are three-fold. First, we propose the community to consider a new, perhaps closer to `infant learning' setting of Few-Shot Learning with Multiple and Complex Semantics (FSL-MCS). Second, in this context we propose a new benchmark for FSL-MCS, and an associated training and evaluation protocol. Third, we propose a new multi-branch network architecture that provides the first batch of encouraging results for the proposed FSL-MCS setting benchmark.

\section{Related Work}\label{sec:related}

\subsection{Few-Shot Learning}
The major approaches to few-shot learning include: metric learning, meta learning (or learning-to-learn), and generative (or augmentation) based methods.

\textbf{Few-shot learning by metric learning:} this type of methods \cite{Snell2017,Rippel2015} learn a non-linear embedding into a metric space where $L_2$ nearest neighbor (or similar) approach is used to classify instances of new categories according to their proximity to the few labeled training examples. Additional proposed variants include \cite{Garcia2017} that uses a metric learning method based on graph neural networks, that goes beyond the $L_2$ metric. Similarly, \cite{Santoro2016,relationnet} introduce metric learning methods where the similarity is computed by an implicit learned function rather than via the $L_2$ metric over an embedding space.

The embedding space based metric-learning approaches are either posed as a general discriminative distance metric learning \cite{Rippel2015}, or optimized on the few-shot tasks \cite{Snell2017,Garcia2017}, via the meta-learning paradigm that will be described next. These approaches show a great promise, and in some cases are able to learn embedding spaces with quite meaningful semantics embedded in the metric \cite{Rippel2015}. The higher end of the performance spectrum for the metric learning based approaches has been achieved when combining these approaches with some additional semantic information. In \cite{caml} class conditioned embedding is used, and in \cite{am3} the visual prototypes are refined using a corresponding label embedding.

\textbf{Few-shot meta-learning (learning-to-learn):} These methods are trained on a set of few-shot tasks (also known as 'episodes') instead of a set of object instances, with the motivation to learn a learning strategy that will allow effective adaptation to new such (few-shot) tasks using one or few examples. An important sub-category of meta learning methods is metric-meta-learning, combining metric learning as explained above with task-based (episodic) training of meta-learning. In Matching Networks \cite{Vinyals2016}, a non-parametric $k$-NN classifier is meta-learned such that for each few-shot task the learned model generates an adaptive embedding space for which the task can be better solved. In \cite{Snell2017} the metric (embedding) space is optimized such that in the resulting space different categories form compact and well separated uni-modal distributions around the category 'prototypes' (centers of the category modes). Another family of meta learning approaches is the so-called 'gradient based approaches', that try to maximize the 'adaptability', or speed of convergence, of the networks they train to new (few-shot) tasks (usually assuming an SGD optimizer). In other words, the meta-learned classifiers are optimized to be easily fine-tuned on new few-shot tasks using small training data. The first of these approaches is MAML \cite{Finn2017} that due to its universality was later extended through many works such as, Meta-SGD \cite{Li2017}, DEML+Meta-SGD \cite{Zhou2018}, and Meta-Learn LSTM \cite{Ravi2017}.
In MetAdapt \cite{doveh2019metadapt} a model is trained to predict the optimal classifier architecture and adapt it at test time to the given task.
In LEO \cite{leo} a MAML like loss is applied not directly on the model parameters, but rather on a latent representation encoding them.

\textbf{Generative and augmentation-based few-shot:} This family of approaches refers to methods that (learn to) generate more samples from the one or a few examples available for training in a given few-shot learning task. These methods include synthesizing new data from few examples using a generative model, or using external data for obtaining additional examples that facilitate learning on a given few shot task. These approaches include: (i) semi-supervised approaches using additional unlabeled data \cite{Fu2015}; (ii) fine tuning from pre-trained models \cite{Wang2016}; (iii) applying domain transfer by borrowing examples from relevant categories \cite{Lim2012} or using semantic vocabularies \cite{Ba2015}; (iv) rendering synthetic examples \cite{Park2015}; (v) augmenting the training examples using geometric and photometric transformations \cite{Krizhevsky2012} or learning adaptive augmentation strategies \cite{Guu2017}; (vi) example synthesis using Generative Adversarial Networks. In \cite{Hariharan2017,Schwartz2018} additional examples are synthesized via extracting, encoding, and transferring to the novel category instances, of the intra-class relations between pairs of instances of reference categories.
Notably, in \cite{Chen2018,Yu2017} label and attribute semantics are used as additional information for training an example synthesis network.

\subsection{Zero-Shot Learning}
A closely related task is zero-shot learning (ZSL). In the ZSL setting only the semantics information is given for the unseen classes. Usually, in ZSL the relation between semantic representations and visual representations are model. Either by mapping semantics to visual \cite{zhang2017learning}, visual to semantics \cite{frome2013devise}, or mapping both to a common embedding space \cite{zhang2015zero}. Other methods use generative models for generating visual samples given the given semantics, e.g. \cite{chen2018zero}, these generated samples can be used to train a classifier.
	
%
%

%
\section{Method}\label{sec:method}
Our approach builds upon the work of \cite{am3}. Our general model architecture is summarized in Figure \ref{fig:model}. It has been shown recently that pre-training the backbone in a standard supervised fashion is good for performance, i.e. simple linear classifier with a softmax non-linearity predicting probabilities for all training categories (no episodic training), for example \cite{wang2019simpleshot}. For this reason we split the training to two phases. In the first phase we perform the fully supervised training of the CNN backbone on the training categories (training procedure similar to \cite{wang2019simpleshot}). At the second phase, the last layer (linear classifier) is discarded and replaced by 2-layer MLP and all previous backbone layers are freezed, and we add the semantics branches. The full model is then trained using the episode-based meta-learning approach proposed by \cite{Vinyals2016}. The training is performed on few-shot tasks (episodes) comprised of one or few image examples for each of the task categories (the so-called support set), as well as one or several query images belonging to these categories (the so-called query set). Each task is simulating a few-shot learning problem. In addition, for our multiple semantics approach, each task is accompanied by semantic information (label and/or description sentence and/or attributes) for each of the task categories. For the labels we use the GloVe embedding \cite{glove} and for descriptions the BERT embedding \cite{bert}, as we observed GloVe performs better for words and BERT for sentences.  

The model is comprised of a visual information branch supported by a CNN backbone computing features both for the training images of the few-shot task and for the query images. As in Proto-Nets \cite{Snell2017}, the feature vectors for each set of the task category support examples are averaged to form a visual prototype feature vector $V$ for that category. In addition, the model contains one or more "semantic branches" for learning to incorporate the additional semantic information. Each semantic branch starts with a pre-trained word or sentence embedding feature extractor (or just a vector in case of attributes), followed by an Multi Layer Perceptron (MLP) generating a "semantic prototypes" $S_i$ to be combined with the corresponding (same category) visual prototype.
For the sake of this combination, each semantic branch is equipped with an MLP calculating "semantic attention" - a coefficient $\alpha_i$ of the semantic prototype of the branch in the overall convex combination of the category prototypes. For computing $\alpha_i$, the attention MLP for each branch can be set to receive as an input one of the task category prototypes generated by either the visual or the semantic branches. We examine the effect of different inputs to the attention MLP of different semantic branches in the ablation study section \ref{sec:ablation} below.

Optionally, our model also allows for adding into the convex combination additional branches with visual prototypes $V$ attended by either of the $S_i$ or $V$ itself. Finally, our model features a task specific cross-entropy loss on the prototype resulting from each (semantic or visual) branch which allows for providing intermediate level supervision for each branch output using the ground truth labels associated to the few-shot tasks (episodes) used for meta-training. These losses admit the softmax normalized logits computed as negative distances between the task query samples and the prototypes produced by each respective semantic (or visual) branch.

To summarize, for each task category, each semantic branch is uniquely determined by its two inputs - the semantic information being processed into the semantic prototype $S_i$ (category label, category description, or attributes vector), and the prototype (visual or semantic) being processed into the semantic attention coefficient $\alpha_i$. The final prototype $P$ for a category in a given few shot task with an associated visual prototype $V$ and semantic prototypes $\{S_1,...,S_k\}$ is computed as:
\begin{equation}
    P = V \cdot \prod_{i=1}^{k} \alpha_i + \sum_{i=1}^{k} \left[ S_i \cdot (1 - \alpha_i) \cdot \prod_{j=i+1}^{k} \alpha_j \right]
    \label{eq:P}
\end{equation}
(please see Fig \ref{fig:model} for the intuitive visualization of Eq. \ref{eq:P}). The final category prototype $P$ is then compared to the query visual feature vector $Q$ (produced by the CNN backbone) for computing the category probability as $prob(Q,P)=SM(-||P-Q||^2)$, where $SM$ stands for the softmax normalization operator. 

Assuming the correct category for the query $Q$ has visual prototype $V$ and semantic prototypes $\{S_1,...,S_k\}$, than the final training loss incorporating the CE losses for all the visual and semantic branches can be written as:
\begin{equation}
    Loss = -\log(prob(Q,V)) + \sum_{r=1}^{k} -log(prob(Q,P_r))
    \label{eq:loss}
\end{equation}
where $P_r$ is the output of the partial computation of equation \ref{eq:P} up until the semantic branch \#$r$:
\begin{equation}
    P_r = V \cdot \prod_{i=1}^{r} \alpha_i + \sum_{i=1}^{r} \left[ S_i \cdot (1 - \alpha_i) \cdot \prod_{j=i+1}^{r} \alpha_j \right]
    \label{eq:P_r}
\end{equation}

\subsection{Implementation details}\label{sec:details}
The CNN backbone is based on the implementation of \cite{wang2019simpleshot} and trained using their provided code and hyper-parameters. The full model combining visual features with multiple semantics is built on top and extends the code provided by the authors of \cite{am3}, also keeping their hyper-parameter setting. The main differences compared to \cite{am3} are: the addition of multiple semantic branches, allowing their cross-conditioning and their cascaded merge scheme; adding intermediate losses after each branch; and pre-training the base layers of the CNN backbone without meta-training as a regular multi-class classifier to all the base classes (and freezing the backbone after pre-training). Our code will be made available upon acceptance.

In our experiments we use the DenseNet-121 backbone CNN \cite{huang2017densely}. After pre-training the linear classifier is replaced by MLP with 512-features hidden and output layers. For each semantic branch, the semantic backbone is a two-layer MLP with a 300-sized hidden layer, and 512-sized output layer. The semantic attention for each branch is a two-layer MLP with a 300-sized hidden layer and a scalar output layer followed by a sigmoid (to normalize the coefficients into a $[0,1]$ range). All semantic branches MLPs include a dropout layer with $0.7$ rate between the hidden layer and the output layer.

After the pre-training of the CNN backbone, all other layers (visual and semantics branches) are trained jointly using the per branch Cross Entropy losses (applied to the predicted logits after a softmax for each branch).
The training is performed using only the training subset of the categories of the few-shot dataset. All parameters are randomly initialized (random normal initialization for the weights and a constant zero for the biases). The category descriptions \textit{mini}ImageNet are obtained automatically from WordNet (see examples in Table \ref{tab:descriptions}). For CUB we sampled 10 random descriptions per category from those used in \cite{xu2018attngan} and used their mean as the category embedding.

\begin{table}
\begin{center}
\small
\begin{tabular}{ll}
\toprule
Label & Description\\
\midrule
\textbf{ImageNet} \\
 Sorrel & \small A horse of a brownish orange color to light \\
 &\small brown color\\
 Consomme &\small  Clear soup usually of beef or veal or chicken \\
Violin & Bowed stringed instrument; this instrument has \\
& four strings and is played with a bow \\
\textbf{CUB} \\
Artic Tern &\small This bird has a long narrow red bill and black crown\\
Geococcyx &\small A tall bird with a long bill and stripped body colors\\
Ovenbird &\small This cute, colorful little bird has almost a cheetah \\
&\small print on its belly and breasts \\
\bottomrule
\end{tabular}
\end{center}
\caption{Examples of descriptions for the \textit{mini}ImageNet and CUB categories. 
}\label{tab:descriptions}
\end{table}

\section{Experimental results}\label{sec:results}

\begin{table}[!htb]
\small
\begin{center}
\begin{tabular}{lcccc}
\toprule
 & \multicolumn{2}{c}{\textit{mini}ImageNet} & \multicolumn{2}{c}{CUB}\\
Method & 1-shot & 5-shot & 1-shot & 5-shot \\
\midrule
\multicolumn{3}{l}{\textbf{Human performance}}\\
\addlinespace
4.5 years old & 70.0 & -- & -- & -- \\
Adult & 99.0 & -- & -- & -- \\
\midrule
\multicolumn{3}{l}{\textbf{No semantics}}\\
\addlinespace
DEML+Meta-SGD \cite{Zhou2018} & $58.5$ & $71.3$ & $66.9$ & $77.1 $\\
CAML \cite{caml} & $59.2 $ & $72.4 $ & -- & --\\
Dist. ensemble \cite{Dvornik2019} 
                  & $59.3$ & $76.9 $
                  & $68.7 $ & $83.5 $ \\  
$\Delta$-encoder \cite{Schwartz2018} & $59.9 $ & $69.7 $ & $69.8$ & $82.6$ \\
LEO \cite{leo} & $61.8 $ & $ 77.6 $ & -- & --\\
MetaOptNet \cite{Lee2019} & $62.6 $ & $78.6 $ & -- & -- \\
MetAdapt \cite{doveh2019metadapt} & 62.8 & 79.3 & -- & --\\
SimpleShot \cite{wang2019simpleshot} & $64.2 $ & $81.5 $ & $70.9 ^*$ & $81.3  ^*$\\
\midrule
\multicolumn{3}{l}{\textbf{With semantics}}\\
\addlinespace
TriNet \cite{Chen2018} & $58.1 $ &  $76.9 $ & $69.6 $ & $\bf 84.1 $\\
AM3-TADAM \cite{am3} & $65.3 $ & $78.1 $ & -- & --\\
Multiple-Semantics (ours) & $\bf 67.3 $ & $\bf 82.1 $ & $\bf 76.1 $ & $82.9 $ \\
\bottomrule
\end{tabular}
\end{center}
\caption{Results for the 5-way \textit{mini}ImageNet and CUB benchmarks. For \textit{mini}ImageNet the semantics used are category labels and textual descriptions; for CUB they are category labels, textual descriptions and attributes. For 1-shot, we observe significant improvement by using multiple semantics. For 5-shot, since the visual information is more reliable semantic information is not as helpful (as observed in previous works). For context we also report human performance of one of the authors and his daughter. The adult performance is very high mainly due to prior familiarity with the categories in question. (* Our run of the code released by the authors)
}\label{tab:results_mini}
\end{table}

\begin{table*}[!htb]
\begin{center}
\begin{tabular}{llcccccr}
\toprule
 & & Branch  & Branch  & Branch  & Branch  & Branch  & \\
& Description &  1 &  2 &  3 &  4 & losses & Accuracy\\
\midrule
a. & Only visual (SimpleShot \cite{wang2019simpleshot}) & - & -  & - & - & - & 64.2 \\
b. & Adding label & l / l & -  & - & - & - & 65.1 \\
c. & Switching to description & d / d & - & - & - & - & 65.4 \\
d. & Ensemble effect (same semantics) & l / l & l / l & - & - & - & 65.3\\
e. & Multiple semantics (2 branches) & l / l & d / v & - & - & - & 65.9\\
f. & Multiple semantics (2 branches) & l / l & d / v & - & - & \checkmark & 66.5\\
g. & Multiple semantics (2 branches) & l / l & d / d & - & - & \checkmark & 66.4\\
h. & Multiple semantics (3 branches) & l / l & d / v & d / d & - & \checkmark & 67.1\\
i. & Multiple semantics (3 branches) & l / l & d / v & d / l & - & \checkmark & 67.3\\
j. & Multiple semantics (4 branches) & l / l & d / v & d / l & v / l & \checkmark & 67.0\\
\bottomrule
\end{tabular}
\end{center}
\caption{
Ablation study performed on 1-shot \textit{mini}ImageNet benchmark. With l = category label, d = category description, v = visual prototype. x/y (e.g. l/l) means x is the branch input and the convex combination parameter is conditioned on y. The 'Branch losses' column marks models that utilize the internal supervision of per branch task CE losses. 
a. Is the SimpleShot baseline. 
b. Is based on AM3 (but different backbone and training procedure).
c. Using only description semantics we observe similar results as when using only labels.
d. The effect of `ensemble', i.e. adding another branch with no extra semantic, is minor ($+0.2\%$ over b.).
e. Adding a second branch with extra semantics adds $0.8\%$.
f-j. Utilizing branch losses with extra semantics adds another $0.6\%$. 
h-i. Third branch adds another $0.8\%$. 
j. Adding a forth branch does not help.}
\label{tab:ablation}
\end{table*}

We have evaluated our approach on the challenging few-shot benchmark of \textit{mini}ImageNet \cite{Vinyals2016} used for evaluation by most (if not all) the few-shot learning works. We also evaluated on the CUB dataset \cite{Welinder2010} which includes another form of semantics, the attributes vector.

\subsection{Datasets}\label{sec:mini}
\textbf{\textit{mini}ImageNet} \cite{Vinyals2016} is a subset of the ImageNet dataset \cite{imagenet}. It contains $100$ randomly sampled categories, each with $600$ images of size $84\times84$. We have used the standard evaluation protocol of \cite{Ravi2017} and evaluated the 1-shot and the 5-shot performance of our method in a 5-way scenario (that is having 1 or 5 training examples for each of the 5 categories in the support set), using 64 categories for training, 16 for validation, and 20 for test. For testing we used 1000 random test episodes (sampled from the test categories unseen during training). The semantics used for each category were the labels and descriptions of the category label in WordNet. 

\textbf{Caltech Birds} (CUB) \cite{Welinder2010} is a dataset consists of $11,788$ images of birds of $200$ species. We use the standard train, validation, and test splits, which were created by randomly splitting the $200$ species into $100$ for training, $50$ for validation, and $50$ for testing. All images are downsampled to $84 \times 84$. For testing we used 1000 random test episodes (sampled from the test categories unseen during training). The semantics used for each category were the labels, attributes (from the original CUB dataset), and descriptions of the categories from \cite{xu2018attngan}.

\subsection{Performance evaluation}
Table \ref{tab:results_mini} summarizes the results of our approach applied to \textit{mini}ImageNet and CUB, compared to the state-of-the-art results with and without using semantics. For both \textit{mini}ImageNet and CUB we used 3 semantic branches in our model. For CUB, the first branch input is label embedding, the second is attribute, and the third is description, the input to all branches' attention module is the visual features. For \textit{mini}ImageNet, the branches configuration used is discussed and analyzed in Section \ref{sec:ablation}.

As can be seen, in the most challenging 1-shot scenario, our multiple semantics based model improves the best previously reported result by $2.3\%$ and $5.2\%$ for \textit{mini}ImageNet and CUB, respectively.
The highest result is achieved using multiple semantic branches receiving as inputs: category labels, more complex (than category labels) description based semantics, and for CUB also attributes. Please see section \ref{sec:ablation} for the description of the branches used to achieve the best result and for the examination of the different branch configurations alternatives.

As expected, the most significant gain from using multiple additional semantics comes when the fewest amount of training examples is available, that is in the 1-shot case. For the 5-shot scenario, when more image examples become available for the novel categories and the visual features are more reliable, the importance of semantic information is decreasing.
Yet, even in this case, the usage of semantics provide some improvement over using only the visual features (SimpleShot \cite{wang2019simpleshot}), $+0.6\%$ for \textit{mini}ImageNet and $+1.6\%$ for CUB.

\subsection{Ablation study}\label{sec:ablation}

\subsubsection{Semantics Effect}
Table \ref{tab:ablation} summarizes the performance of different (multiple) semantic branch configuration alternatives and other aspects of the proposed approach evaluated using the 1-shot \textit{mini}ImageNet test.

Combining visual features with labels provides $0.9\%$ improvement over the SimpleShot \cite{wang2019simpleshot} baseline, this result supports the results reported in \cite{am3} (Table \ref{tab:ablation}a-b). As can be seen from the table, using the more complex description semantics instead of the labels used in \cite{am3} does not by itself improve the performance by much, only $+0.3\%$ (Table \ref{tab:ablation}c). Next we tested the effect of `semantic ensemble' by using multiple semantic branches relying only on the labels, without adding additional semantic information (descriptions). This leads to only a slight improvement over a single branch, $+0.2\%$ (Table \ref{tab:ablation}d).

A more significant improvement of $0.8\%$ over using just labels baseline is attained by incorporating additional semantic information (descriptions conditioned on the labels) in the second semantic branch (Table \ref{tab:ablation}e). Introducing intermediate supervision in the form of per branch task specific Cross Entropy losses brings even more significant improvement of $1.4\%$ over the baseline (Table \ref{tab:ablation}f) underlining the importance of this component.

In further tests, all using the intermediate per branch loss, we see that conditioning the second (description) branch on itself does not bare improvements (Table \ref{tab:ablation}g), yet a substantial improvement of $+2\%$ over the baseline is obtained when adding the the self-attending description as the third semantic branch (Table \ref{tab:ablation}h). Changing the third semantic branch to use labels for attending to the added description semantics, and thus utilizing the most comprehensive conditioning strategy (attending using all prior inputs to the combination) leads to the maximal $2.2\%$ improvement over the baseline (Table \ref{tab:ablation}i) and comprises our final method for \textit{mini}ImageNet. Finally, in additional experiments we have observed that adding additional semantic branches, while re-using the same semantic information, does not help the performance (Table \ref{tab:ablation}j, as an example). This is intuitive as this likely leads to increased over-fitting due to adding more trainable network parameters.

\subsubsection{Backbone Architecture}
We also tested the effect of the backbone architecture on the performance of our approach. Table \ref{tab:ablation_arch} presents the comparison between the relatively small ResNet-10 architecture and the larger DenseNet-121 architecture. Moving to a stronger backbone we observed substantial improvement for 5-shot but no improvement for 1-shot. It supports our observation that for the 1-shot case the model relies more heavily on the semantics branches while for 5-shot the visual features are more important.

\begin{table}[!htb]
\begin{center}
\small
\begin{tabular}{lcc}
\toprule
Architecture & 1-shot & 5-shot\\
\midrule
ResNet-10 & 67.2 & 74.8 \\
DenseNet-121 & 67.3 & 82.1 \\
\bottomrule
\end{tabular}
\end{center}
\caption{Effect of architecture on performance for \textit{mini}ImageNet}
\label{tab:ablation_arch}
\end{table}	
	
\section{Summary \& conclusions}\label{sec:summary}

In this work, we have proposed an extended approach for few-shot learning with additional semantic information. We suggest making few-shot learning with semantics closer to the setting used by human infants: we build on multiple semantic explanations (name, attributes and description) that accompany the few image examples and utilize more complex natural language based semantics rather than just the name of the category. In our experiments, we only touch the tip of the iceberg of the possible approaches for using descriptive and multiple semantics for few-shot learning. Many other ways for combining multiple semantic information with visual inputs are possible and are very interesting topics for the follow-up works. In particular, we offer to investigate the following possible future work directions:
\begin{itemize}
\itemsep0em 
    \item Using instance level instead of category level semantics.
    \item Attending to visual and semantic branches combining information from all the task categories. In the current experiments, the coefficient of each category semantic prototype is computed from the attention MLP input of the corresponding category (either semantic or visual prototype of the same category). A future work may learn to attend based on the entire task jointly.
    \item Alternative non-linear (e.g. MLP) combination schemes for visual and semantic prototypes instead of the (linear) convex combination we use here.
    \item Learning alternative metrics, conditioned on the semantics, for comparing prototypes and query image features (e.g. learned Mahalanobis distance, with covariance matrix computed from semantic prototypes).
    \item Semantic ensembles: instead of combining prototypes, combine logits resulting from different semantic and visual branches.
    \item Further exploring different semantic sources and prototype / attention combinations. E.g. using the categories hierarchy \cite{akata2016label} or investigating into multi-modal sources of semantics, such as audio / smell / touch / taste, to further approximate the human infant learning environment.
\end{itemize}

{\small
\bibliographystyle{ieee}
\bibliography{main}
}

\end{document}